\documentclass[letterpaper, 10 pt, conference]{ieeeconf}  %

\IEEEoverridecommandlockouts                              %

\usepackage{amsmath} %
\usepackage{amssymb}  %

\usepackage{subcaption}
\usepackage[nolist]{acronym}

\usepackage[inkscapelatex=false]{svg}
\usepackage{booktabs} %
\usepackage{multirow}
\usepackage{array}
\usepackage{tabularx}
\usepackage{xfrac}
\usepackage{placeins}
\usepackage{comment}
\usepackage{cite}

\DeclareCaptionFont{eightpt}{\fontsize{8}{10}\selectfont}
\usepackage[table]{xcolor} %

\usepackage{hyperref}

\newcommand{\red}[1]{\textcolor{red}{#1}}

\title{\LARGE \bf
CrowdQuery: Density-Guided Query Module for Enhanced 2D and 3D Detection in Crowded Scenes
}

\author{Marius Dähling$^{1, 2}$ \and Sebastian Krebs$^{2, 3}$ \and J. Marius Zöllner$^{1, 4}$%
\thanks{$^{1}$Karlsruhe Institute of Technology (KIT), Karlsruhe, Germany}%
\thanks{$^{2}$Mercedes-Benz AG, Research and Development, Stuttgart, Germany}%
\thanks{$^{3}$Intelligent Vehicles Group at TU Delft, Delft, Netherlands}%
\thanks{$^{4}$Research Center for Information Technology (FZI), Karlsruhe, Germany}%
\thanks{Contact: marius.daehling@mercedes-benz.com}%
}

\begin{acronym}
\acro{IoU}{Intersection-over-Union}
\acro{NMS}{Non-Maximum Suppression}
\acro{ADAS}{advanced driver-assistance systems}
\acro{AP}{Average Precision}
\acro{mAP}{mean Average Precision}
\acro{AR}{Average Recall}
\acro{CQ}{CrowdQuery}
\acro{FoV}{Field of View}
\end{acronym}

\begin{document}
\bibliographystyle{citation/IEEEtran}

\maketitle

\begin{abstract}
This paper introduces a novel method for end-to-end crowd detection that leverages object density information to enhance existing transformer-based detectors. 
We present \ac{CQ}, whose core component is our \ac{CQ} module that predicts and subsequently embeds an object density map. 
The embedded density information is then systematically integrated into the decoder.
Existing density map definitions typically depend on head positions or object-based spatial statistics.
Our method extends these definitions to include individual bounding box dimensions. 
By incorporating density information into object queries, our method utilizes density-guided queries to improve detection in crowded scenes. 
CQ is universally applicable to both 2D and 3D detection without requiring additional data. 
Consequently, we are the first to design a method that effectively bridges 2D and 3D detection in crowded environments. 
We demonstrate the integration of CQ into both a general 2D and 3D transformer-based object detector, introducing the architectures CQ2D and CQ3D. 
CQ is not limited to the specific transformer models we selected. 
Experiments on the STCrowd dataset for both 2D and 3D domains show significant performance improvements compared to the base models, outperforming most state-of-the-art methods. 
When integrated into a state-of-the-art crowd detector, \ac{CQ} can further improve performance on the challenging CrowdHuman dataset, demonstrating its generalizability.
The code is released at \url{https://github.com/mdaehl/CrowdQuery}.
\end{abstract}

\section{INTRODUCTION}
Pedestrian detection is a critical component in addressing navigation challenges for robots \cite{mateusEfficientRobustPedestrian2018} and ensuring the safety of autonomous driving systems \cite{yangRealtimePedestrianVehicle2018}. 
Upcoming advanced driver-assistance systems (ADAS) increasingly focus on urban environments, where pedestrians often appear as crowds, resulting in heavy occlusions. 
In these applications, it is essential to detect all instances, as the failure to recognize a single pedestrian who suddenly crosses the street can result in severe consequences. 
Conversely, false positives can pose risks, such as unnecessary emergency braking, which can lead to additional hazards. 
In general, this problem is referred to as \textbf{crowd detection} and presents significant challenges.

General object detection is primarily dominated by transformer models, after having demonstrated impressive results \cite{sunWhatMakesEndtoEnd2021, zhuDeformableDETRDeformable2021}. 
However, when applying transformers directly to crowd detection, they often face difficulties. 
According to the work of \cite{linDETRCrowdPedestrian2021}, one of the main causes is that initially uniformly distributed queries are expected to adapt to local dense regions in the image during training. 
Although the current focus in crowd detection is predominantly on 2D detection, object detection in the ADAS domain is progressively developing towards 3D detection. 
While there exist many 3D datasets \cite{caesarNuScenesMultimodalDataset2020, geigerAreWeReady2012, congSTCrowdMultimodalDataset2022}, it remains unclear to what extent 3D detectors are affected by these difficulties, as few of the datasets are dedicated to crowds. 
As a result, it is still difficult to bridge the gap between 2D and 3D detection in this domain. 
In this paper, we focus on transformer methods and improved crowd detection for 2D and 3D approaches.

\begin{figure}[t]
    \centering
    \begin{subfigure}[b]{0.49\linewidth}
        \centering
        \includegraphics[width=\linewidth]{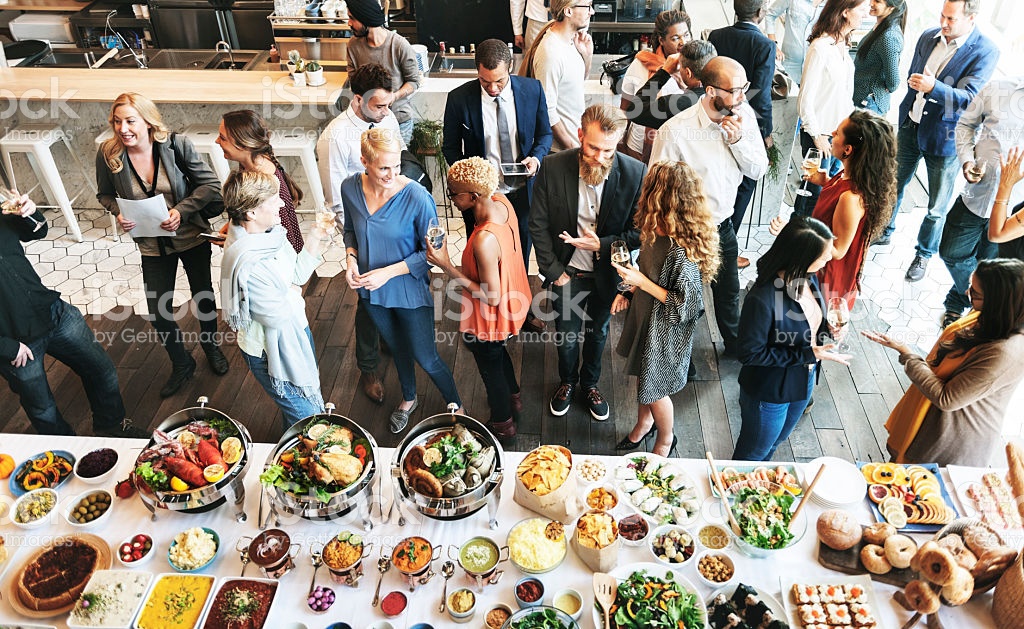}
        \caption{Original Image}
        \label{fig:orig_img}
    \end{subfigure}
    \hfill
    \begin{subfigure}[b]{0.49\linewidth}
        \centering
        \includegraphics[width=\linewidth]{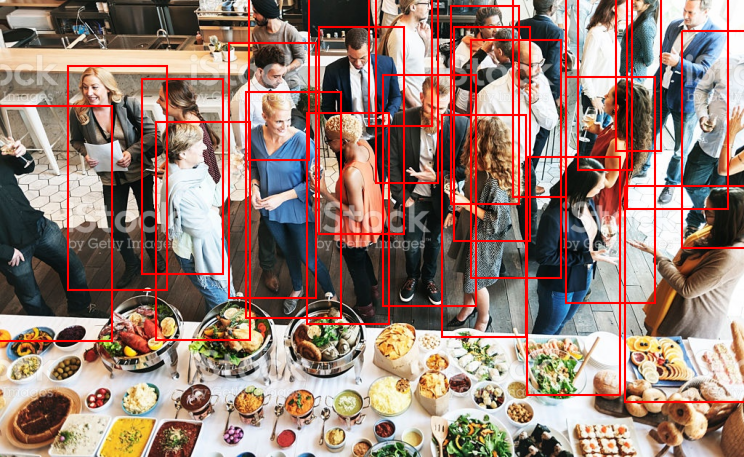}
        \caption{Annotated Image}
        \label{fig:bbox_img}
    \end{subfigure}
    \vfill
    \begin{subfigure}[b]{0.49\linewidth}
        \centering
        \includegraphics[width=\linewidth]{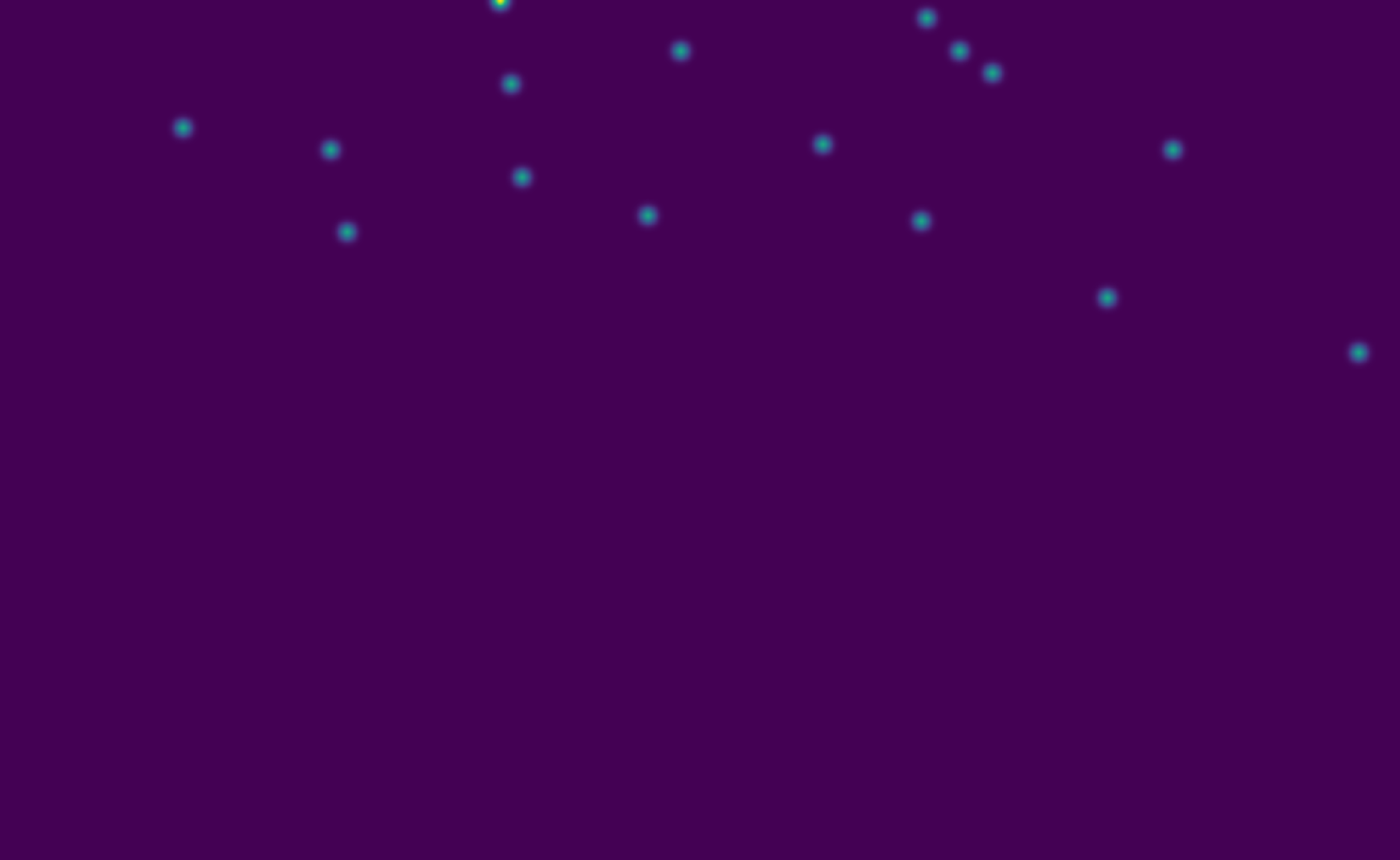}
        \caption{Current Density Map}
        \label{fig:ddad_density}
    \end{subfigure}
    \hfill
    \begin{subfigure}[b]{0.49\linewidth}
        \centering
        \includegraphics[width=\linewidth]{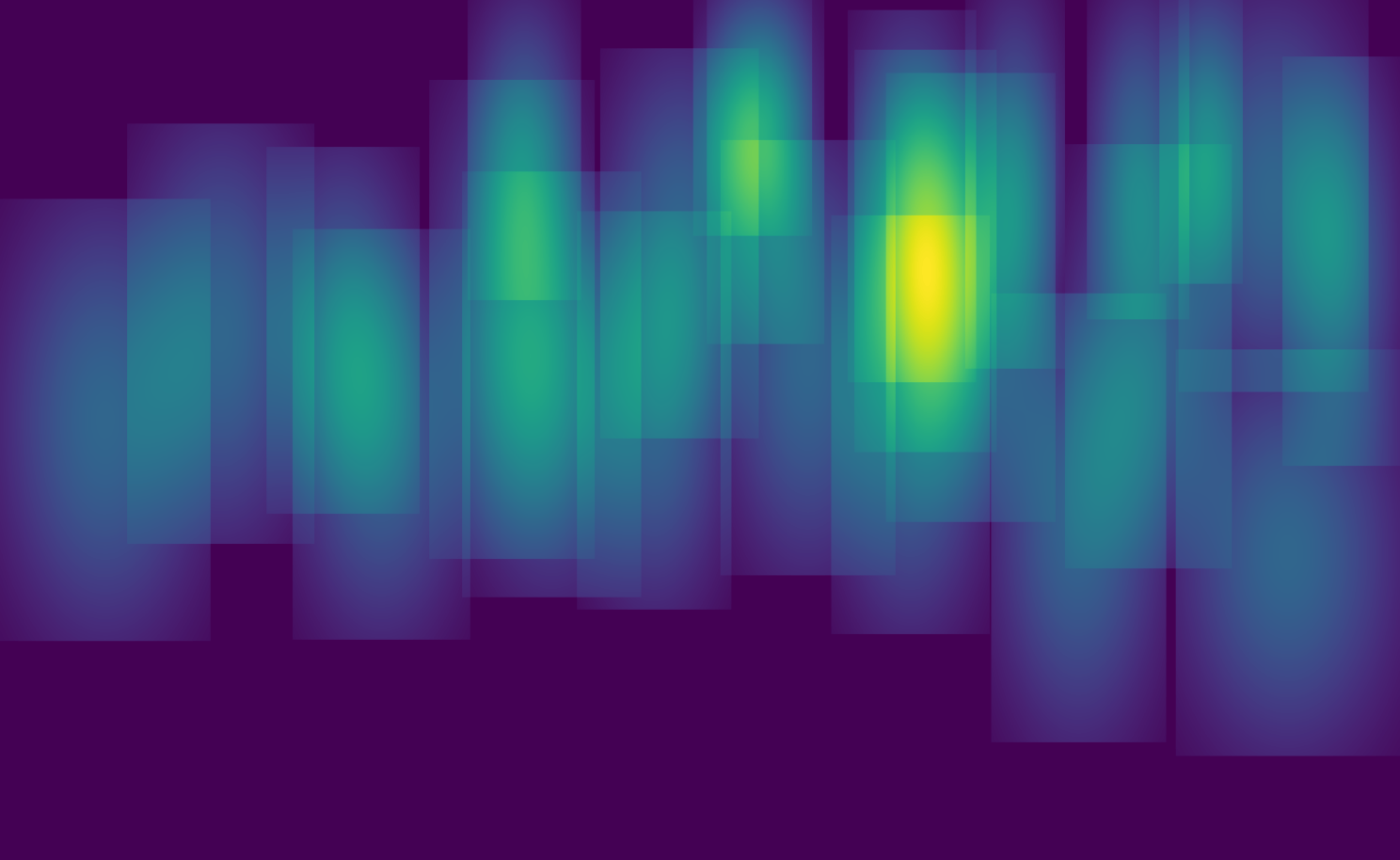}
        \caption{Our Density Map}
        \label{fig:my_density}
    \end{subfigure}
    \caption{Comparison between density formulations. \ref{fig:orig_img} shows a crowded image (taken from \cite{shaoCrowdHumanBenchmarkDetecting2018}) and \ref{fig:bbox_img} visualizes the bounding boxes used in 2D object detection. The second row compares the existing head-based density map \cite{tangDDADDetachableCrowd2022} in \ref{fig:ddad_density} with our novel version in \ref{fig:my_density}.}
    \label{fig:density_types}
\end{figure} 

Optimizing transformers for specific tasks often involves integrating task-related knowledge to guide the object queries. 
Examples include \cite{zhangMonoDETRDepthguidedTransformer}, which uses explicit depth information for 3D detection, and \cite{zhangDecoupledDETRSpatially}, which employs separate query sets for 2D detection to focus on localization and classification. 
Inspired by these works, we propose the integration of density maps, which are commonly used in crowd counting \cite{lempitskyLearningCountObjects}, into the task of pedestrian detection in crowded scenes.
DDAD \cite{tangDDADDetachableCrowd2022} first used density maps to guide a detector, though not in an end-to-end model. 
Despite its simplicity, DDAD's approach was successful but had drawbacks: its density maps treated each object as a single point, ignoring bounding-box dimensions.
They relied on Faster R-CNN \cite{renFasterRCNNRealTime2016}, leaving open the question of how density information can be integrated into transformer models and the extent of its potential benefits.

In this paper, we propose a query-based module, \mbox{\textbf{\acf{CQ}}}, that integrates density map information via cross-attention as query guidance into end-to-end transformer models.
Unlike previous work, our density map is tailored to individual bounding boxes, preserving their spatial information (see Figure \ref{fig:density_types}). 
The lightweight submodule predicts the density map using existing backbone features, which is then embedded and fed into the decoder via cross-attention to guide the detection. 
Our method is universally applicable to both 2D and 3D crowd detection, demonstrating the benefits of addressing crowd detection in general through unified methods.
It requires only a query-based decoder, which, to the best of our knowledge, applies to all transformer-based detectors as of now.
Additionally, this method allows crowd detection to take advantage of the latest object detectors in both domains by integrating our module.
Our main contributions are summarized as follows:
\begin{itemize}
    \item We extend the density map definition by integrating individual bounding box characteristics rather than general statistics or human head positions.
    \item We design a universal module that integrates object density to improve object detection, not just in 2D but also in 3D, without the need for extra data.
    \item To the best of our knowledge, we are the first to evaluate 3D camera-based crowd detection in the pedestrian domain.
    \item We present a method that significantly enhances the performance of general object detectors in the crowd detection task in both 2D and 3D.
\end{itemize}

\section{RELATED WORK}

\textbf{Pedestrian Detection / Crowd Detection:}
In pedestrian and more specifically crowd detection, predominantly three types of detectors are used.
There are keypoint-based \cite{duanCenterNetKeypointTriplets2019, dongCentripetalNetPursuingHighQuality2020, liuGigaHumanDetExploringFullBody2024}, box-based  \cite{renFasterRCNNRealTime2016, caiCascadeRCNNDelving2018, linFocalLossDense2017, hasanGeneralizablePedestrianDetection2021, chuDetectionCrowdedScenes2020},  and query-based detectors \cite{zhang2023dino, zhuDeformableDETRDeformable2021, zhengProgressiveEndtoEndObject2022, zhang2023dense}. 
For each of these detector types, a wide variety of architectures exists. 
Their key difference lies in the representation they utilize to learn to detect objects. 
Box-based detectors directly use bounding boxes, which they optimize during training, whereas keypoint-based detectors look for specific points, such as corner or center points. 
Query-based detectors, on the other hand, rely on a latent representation to learn the task.

Both the box- and keypoint-based detectors usually require \ac{NMS} as a post-processing step when running inference. 
Vanilla \ac{NMS} often faces the issue that it is difficult to tune and its \ac{IoU} criterion easily results in too many objects being suppressed in dense scenarios \cite{songOptimalProposalLearning2023}.
This problem has long been known, and in the meantime, numerous \ac{NMS} variants \cite{bodlaSoftNMSImprovingObject2017, heBoundingBoxRegression2019, liuAdaptiveNMSRefining2019} have been developed to tackle this. Another compelling technique was presented by \cite{chuDetectionCrowdedScenes2020}. 
In their work, they allow each bounding box proposal to predict multiple instances that cannot suppress each other during \ac{NMS}, which helps to detect overlapping objects. 
Recently, the trend is shifting towards end-to-end models. 
An example is provided by \cite{songOptimalProposalLearning2023}, which employs an architecture that continuously reduces the number of possible detections through multiple prediction stages and thus makes \ac{NMS} obsolete.

\begin{figure*}[ht]
    \centering
    \includegraphics[width=\linewidth]{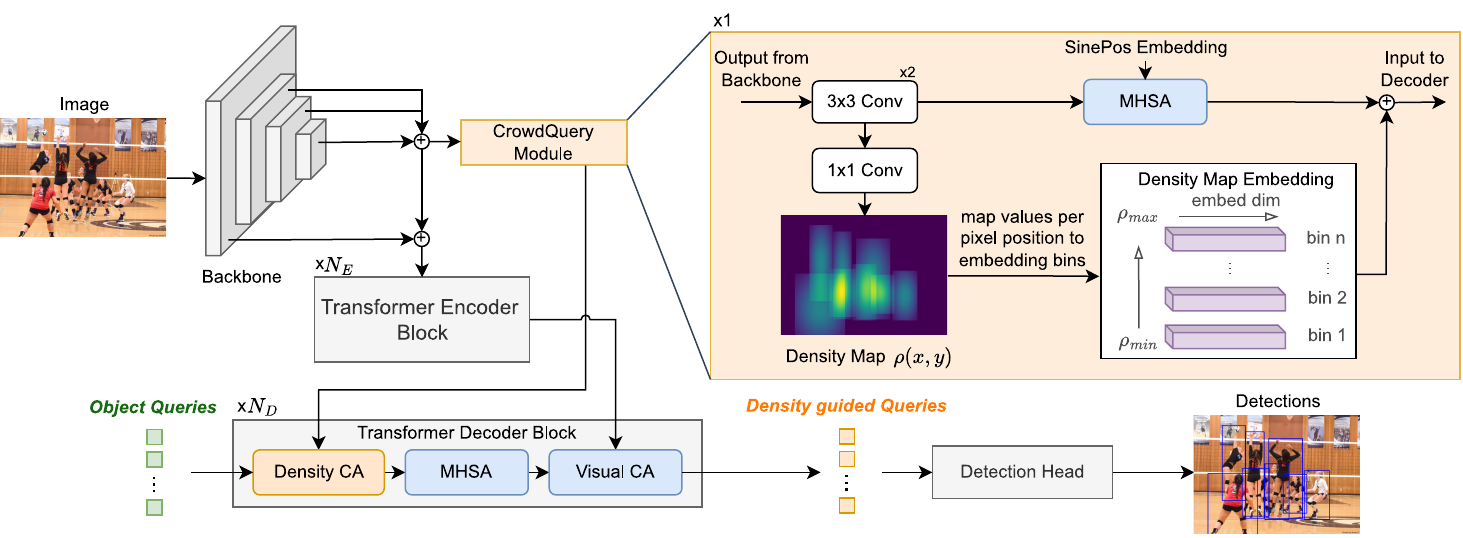}
    \caption{The overall architecture of our proposed method, CrowdQuery, integrated into a query-based 2D object detector architecture for detection. The CrowdQuery module processes the backbone features to predict a density map $\rho(x,y)$, which is embedded and combined with latent density features. Subsequently, the features are fed into the decoder to guide the object queries in the crowd detection task. The number of encoder stages $N_E$ and decoder stages $N_D$ depend on the selected base model. We abbreviate the cross-attention as \textit{CA} and the multi-head self-attention as \textit{MHSA}.}
    \label{fig:base_architecture}
\end{figure*}

Query-based detectors, which we refer to as transformers, are end-to-end by default. 
This is achieved through the training process, which ensures that each ground truth can only be matched with a single prediction. 
Therefore, the model is implicitly trained to suppress duplicates. 
However, as transformers had difficulties in crowded scenarios, \cite{linDETRCrowdPedestrian2021} began adapting them. 
They redefined the self-attention by letting each query only attend to its nearest neighbors and rectified the attention window to avoid it being too narrow. 
The idea of refining the proposals has also proven successful in transformers, as \cite{zhengProgressiveEndtoEndObject2022} showed. 
They divide the queries into accepted and noisy queries based on their score, and use the accepted queries afterwards to refine the noisy queries.

Recently, the work of \cite{zhang2023dense} integrated the concept of \ac{NMS} into query-based detectors. 
Their approach assigns an object query to each position in the feature map, generating a large pool of queries. 
From this set, a subset is selected based on confidence scores. To further refine the selection, class-agnostic \ac{NMS} is applied after each decoder stage, progressively reducing the number of queries. Crucially, the original one-to-one assignment is preserved, and loss is backpropagated only through the selected queries, ensuring that the method remains fully end-to-end.

In the realm of 3D detection, various general detectors have been developed. 
Among these are GUPNet \cite{lu2021geometry} and MonoFlex \cite{zhangObjectsAreDifferent2021} that build on CenterNet \cite{duanCenterNetKeypointTriplets2019} or \mbox{MonoDETR} \cite{zhangMonoDETRDepthguidedTransformer} as an extension of Deformable DETR \cite{zhuDeformableDETRDeformable2021}.

In summary, there is a broad spectrum of 2D object detectors dedicated to crowd detection. 
However, in the 3D domain, the detectors are not yet adapted to dense scenarios. 
This is where our research becomes pivotal, as we engineer a universal module that builds on the existing methods to enable both 2D and 3D end-to-end crowd detection.

\textbf{Object Density:}
Object density has long been used before the rise of neural networks and was introduced by \cite{lempitskyLearningCountObjects}, who applied it to crowd counting. 
In the crowd counting domain, there are usually no bounding box labels available. 
Instead, the labels consist of single points that are placed at the center of people's heads. 
As a result, the labels for a single image $I$ are defined by a set of 2D points \mbox{$\mathbf{P} = \{P_1, ..., P_n\}$}, where $n$ is the total number of objects in the image. 
The ground truth density map is defined as follows:
\begin{equation}
        \forall p \in I, F(p)  = \sum\limits_{P_i \in \mathbf{P}} \mathcal{N}(p|\mu=P_i, \sigma^2\mathbf{I}),
\end{equation}
where the term $\mathcal{N}(p|\mu=P, \sigma^2\mathbf{I})$ denotes a 2D isotropic Gaussian distribution with its mean placed at point $P$  (equal to the object's head center) and a variance $\sigma^2$ evaluated at pixel position $p$.
The work of \cite{lempitskyLearningCountObjects} considers $\sigma$ to be fixed.
\cite{zhangSingleImageCrowdCounting2016} extends this approach by making the Gaussian adaptive, meaning that each object is assigned its own $\sigma$. The value of $\sigma$ is determined using the average distance to the $K$ nearest targets, making it proportional to the local object density.

Later, \cite{liDensityMapGuided2020} adapted the density map definition, more specifically $\sigma$, by incorporating bounding box statistics. Accounting for the presence of different classes, they defined their Gaussian for a single class $j$. The Gaussian calculation is based on the average bounding box height $\bar{h}$ and width $\bar{w}$ of an object category, resulting in the following formula:
\begin{equation}
    \sigma_i=\frac{1}{2}\sqrt{\bar{h}_j^2+\bar{w}_j^2}.
\end{equation}

While their model performs object detection, it does not use the density map within the model itself. The density map is predicted by a separate model, and after applying a predefined threshold, a mask is returned. The mask is used to guide the region proposal of the subsequent object detector.

The first and only explicit use of density map information within a detector is by DDAD \cite{tangDDADDetachableCrowd2022}. They rely on the density map definition of \cite{lempitskyLearningCountObjects} based solely on head position and integrate it into a Faster R-CNN \cite{renFasterRCNNRealTime2016} architecture.

\acresetall
\section{METHOD}
We begin by introducing our novel density map representation and the resulting architecture of our \ac{CQ} module in Section \ref{subsection:crowdquery}.
Following this, we demonstrate in Section \ref{subsection:integration} the integration of \ac{CQ} into 2D and 3D architectures.
Finally, Section \ref{subsection:model_training} discusses how the integration of \ac{CQ} affects model training. The overall architecture, using the example of a 2D detector, is illustrated in Figure \ref{fig:base_architecture}.

Throughout this work, the terms \textit{CrowdQuery} and \textit{CrowdQuery module} are frequently used and may seem identical. However, it is important to note that the \ac{CQ} module itself produces additional features to aid detection, while \ac{CQ} refers to the complete method, which includes the necessary operations to merge these features into the model's decoder.

\subsection{CrowdQuery Module}\label{subsection:crowdquery}
\textbf{Density Map:}
In order to utilize density information in the \ac{CQ} module, the target density map needs to be defined.
The density map $\rho(x,y)$ serves as a mapping function that assigns density values to pixel positions $(x,y)$.

Building on prior work, the Gaussian distribution is adopted as a starting point.
The goal is to find an individual distribution for each object to handle the variety in bounding box shapes. 
Simultaneously, the density distribution must account for asymmetry, as humans and their bounding boxes are typically not square due to natural body proportions.
Considering these requirements, we formulate the following 2D Gaussian distribution at a pixel position $(x,y)$ as:
\begin{equation} \label{eq:gauss}
    \mathcal{N}(x,y)= \frac{1}{2 \pi \sigma_x \sigma_y} e^{-\frac{1}{2} \left( \big(\frac{x}{\sigma_x}\big)^2 + \left( \frac{y}{\sigma_y}\right)^2 \right)}.
\end{equation}
The standard deviations $\sigma_x$ and $\sigma_y$ are defined to be proportional to the bounding box width $w_{\text{bbox}}$ and height $h_{\text{bbox}}$, respectively. 
By introducing the hyperparameter $d$, we can scale them relative to each other as follows:
\begin{equation}
    \sigma_x = d w_{\text{bbox}} \quad \text{and} \quad \sigma_y = d h_{\text{bbox}}.
\end{equation}

\begin{figure}
    \centering
    \includegraphics[width=0.9\linewidth]{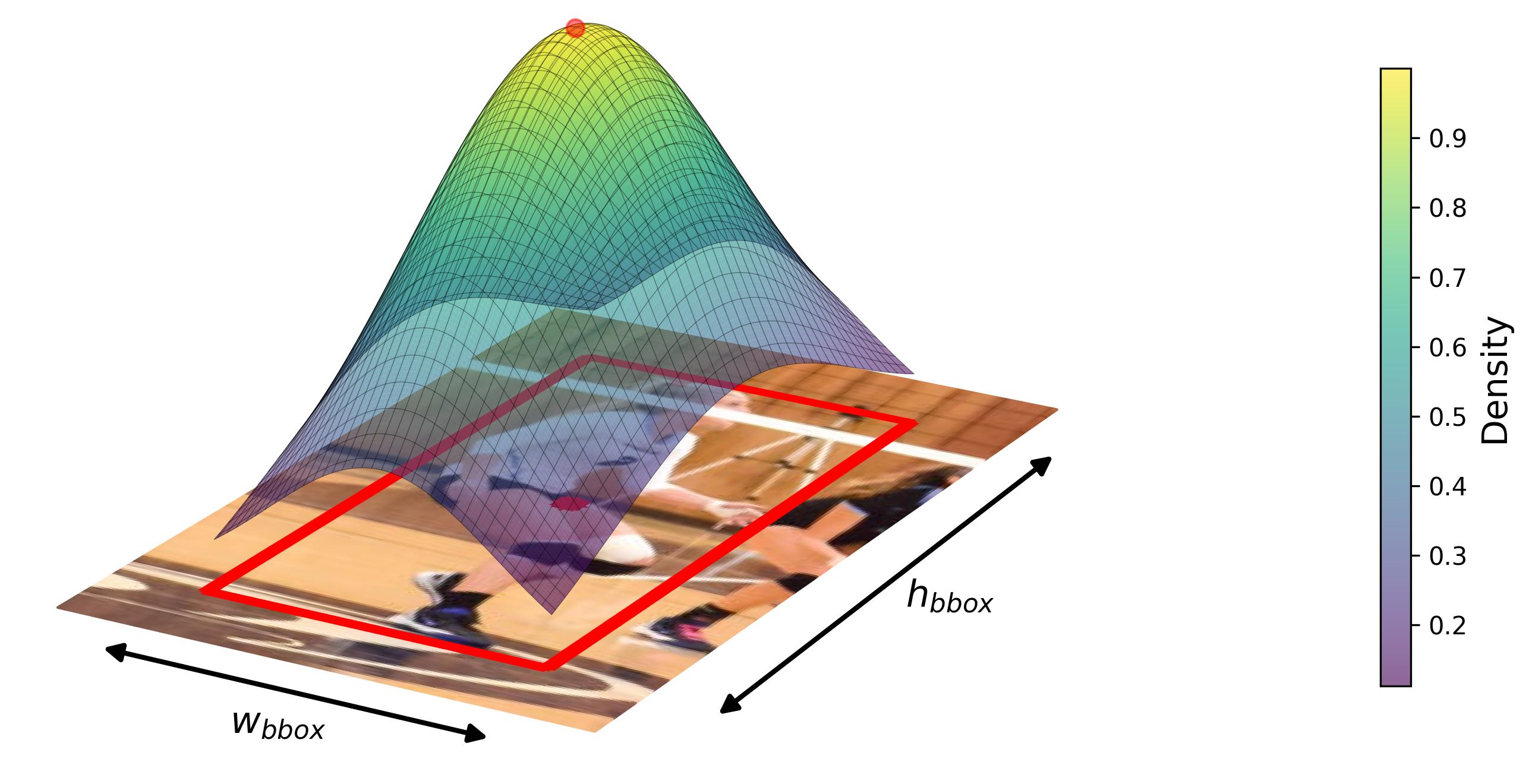}
    \caption{Image crop illustrating a bounding box (red rectangle) and the respective density distribution above. The lower red point marks the bounding box center, and the upper one the projected center in the density distribution.}
    \label{fig:density_dist}
\end{figure}

The formulation in Equation \ref{eq:gauss} takes into account the individual shape and size of the bounding box via $\sigma_x$ and $\sigma_y$. 
However, the normalization constant adjusts the value range, which causes large bounding boxes to contribute little to the density compared to small boxes. 
Removing the normalization constant, we are able to circumvent this effect and obtain an evenly distributed density, regardless of whether the objects are grouped closer or further apart from the camera. 
Gaussian distributions without normalization are also referred to as Gaussian Windows. 
Besides that, we expect our density function to be zero at positions where no bounding box is present. 
Therefore, the density distribution with respect to a single bounding box $B_i$ is defined as:
\begin{equation}
    f_i(x, y) = \begin{cases}
    e^{-\frac{1}{2} \left( \frac{(x - \mu_x)^2}{\sigma_x^2} + \frac{(y - \mu_y)^2}{\sigma_y^2} \right)} & \text{if } (x,y) \in B_i\\
    0  & \text{else },
    \end{cases}    
\end{equation}
where $\mu_x$ and $\mu_y$ correspond to the bounding box center in pixel coordinates. 
We note that for $\sigma \to \infty$ our density map collapses into an occupancy map. 
Figure \ref{fig:density_dist} illustrates the density distribution to better understand the above formulation. 
The figure highlights the adaptation to the non-square box shape as well as the peak value of one in the center. 

The final density map is calculated by summing the partial density maps of all bounding boxes $\mathcal{B}$ in the image:
\begin{equation}
    \rho(x,y)=\sum_{B_i \in \mathcal{B}} f_i(x,y).
\end{equation}

\textbf{Module Architecture:}
Our \ac{CQ} module (see Figure \ref{fig:base_architecture}) consists of two branches that process features independently and combine them in the end before being merged into the decoder. 
The upper branch encodes the density features and ensures global information exchange.
In contrast, the lower branch aims to learn a suitable density map embedding based on the preceding predicted density map.
The density map prediction is a stand-alone component, though its primary goal is model guidance. 
Hence, the actual density map prediction is inferior to finding an ideal density embedding. 

Using ResNet-50 \cite{heDeepResidualLearning2016} as the backbone, the final three feature maps serve as input to our \ac{CQ} module.
These feature maps are merged beforehand to the intermediate scale via up-/downsampling. The module is not limited to this specific backbone, and feature merging may not be necessary if another backbone is selected. 
Using the backbone features as input, no additional image encoder is required. 

The base architecture of our module was introduced by MonoDETR \cite{zhangMonoDETRDepthguidedTransformer} that demonstrated its effectiveness for subtask prediction and encoding. 
In order to receive dedicated density features, two 3x3 convolutions are utilized. 
These features are further processed in two ways. First, as shown in the upper branch of our \ac{CQ} module, the features are enhanced using multi-head self-attention. 
Following the common practice of \cite{carionEndtoEndObjectDetection2020}, we also integrate a SinePosEmbedding, which encodes the position of each pixel using sine and cosine functions. 
Second, the lower branch starts by predicting a density map using a 1x1 convolution. 

We approach density map prediction as a pixel-wise regression problem. 
Although a categorical formulation based on discretized density values allows for direct mapping to respective embeddings, it has a significant drawback: 
When training a model to produce categorical output, each incorrect prediction is treated equally. 
The loss function does not consider the similarity between adjacent classes, or in this context, density value bins.
After predicting the density map, each pixel value is discretized and mapped to a bin of the density map embedding.
Each bin corresponds to a range of density values.
The embeddings used in this process are learned during training.
Following the mapping procedure results in an embedded density map $\rho_{\text{emb}} \in \mathbb{R}^{h_\rho \times w_\rho \times C_H}$, where $C_H$ denotes the base model's number of hidden dimensions. 
The final resolution of the embedded density map ($h_\rho$, $w_\rho$) is identical to that of the input feature map.

Using a total of $n_{\text{bins}}$ uniformly distributed bins, the mapping from density value $\rho(x,y)$ to the $k$-th embedding bin is defined as follows:
\begin{equation}
    k =\left\lfloor \frac{\rho(x,y) - \rho_{\text{max}}}{\frac{\rho_{\text{max}} - \rho_{\text{min}}}{n_{\text{bins}}}} \right\rfloor,
\end{equation}
where $\rho_{\text{max}}$ defines the upper and $\rho_{\text{min}}$ the lower bound density. 
 Values outside this range are clipped.

Finally, the encoded density features (upper branch) are added entry-wise to the embedded density map (lower branch).

\subsection{Integration into 2D/3D Architectures}\label{subsection:integration}
In this subsection, we explain the integration of \ac{CQ} into both a 2D and 3D object detector.
We choose the 2D detector Deformable DETR \cite{zhuDeformableDETRDeformable2021} and the 3D detector MonoDETR \cite{zhangMonoDETRDepthguidedTransformer} as base models because they closely adhere to the general core architecture of DETR \cite{carionEndtoEndObjectDetection2020}.
We focus on changes to the decoder, as the rest of the architecture, including the Deformable DETR encoder and detection heads, remains unchanged. The selected architectures that are extended by \ac{CQ} are referred to as  CQ2D and CQ3D.

\textbf{CQ2D:}
In addition to the standard decoder of Deformable DETR, the density information is incorporated alongside the object queries and image features. 
We decide to integrate the density information prior to the image encoder information to prepare the queries before being exposed to the full image information of the encoder. 
Therefore, as a first step, the cross-attention between the density embedding and the object queries is performed. 
We end up with enhanced object queries, which we refer to as density-guided queries. 

\begin{figure}
    \centering
    \includegraphics[width=\linewidth]{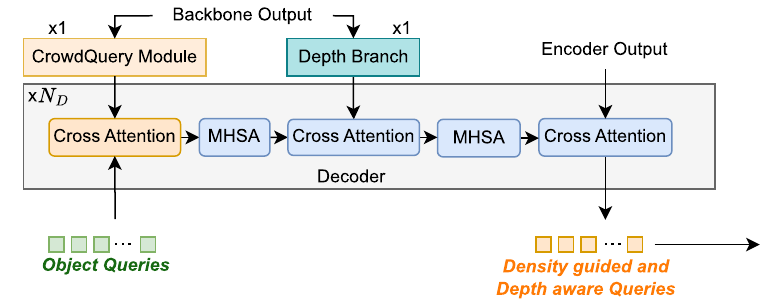}
    \caption{Excerpt of the CQ3D architecture showing the integration of our CrowdQuery module using MonoDETR \cite{zhangMonoDETRDepthguidedTransformer}. We combine their depth components into a depth branch for a simplified visualization. \textit{MHSA} denotes the multi-head self-attention.}
    \label{fig:architecture_3d}
\end{figure}

\textbf{CQ3D:}
Similarly to CQ2D, the input into the CQ3D decoder comprises object queries, encoded image features, and density information. 
Building upon MonoDETR as our base model, we additionally incorporate depth information.
The full integration of our \ac{CQ} module into MonoDETR is illustrated in Figure \ref{fig:architecture_3d}. 
We highlight that the density information is explicitly processed before the depth or any other task-related information.
The justification is that the density prediction is not inherently relevant for the final task. 
Its purpose is the guidance for the detection, whereas the depth and image features are explicitly required to predict 3D bounding boxes. 
Following the cross-attention between the object queries and density features, the subsequent self-attention mechanism enables the exchange of information among the queries.
This is succeeded by MonoDETR's original decoder.
The decoder consists of the density cross-attention, another self-attention layer, and the visual cross-attention, which integrates the image features.

\subsection{Model Training}\label{subsection:model_training}
Integrating \ac{CQ} into different architectures does not alter the overall training procedure. 
The only component affected by this integration is the loss function.

The loss consists of the base model's loss $\mathcal{L}_{base}$ and the added density map loss $\mathcal{L}_{\rho}$. 
We define the density map loss as a pixel-wise L1 loss between the target density map $\rho$ and the predicted density map $\hat{\rho}$. 
Notably, in both the 2D and 3D applications, 2D bounding boxes are used to generate the target density map.
The combined loss for a single image can be formulated as:
\begin{equation}
\begin{aligned}
\mathcal{L}_{\text{total}} &= \mathcal{L}_{\text{base}} + \lambda\mathcal{L}_{\rho} \\
&= \mathcal{L}_{\text{base}} + \lambda \frac{1}{h_{\rho}w_{\rho}}\sum_{x=0}^{h_{\rho}} \sum_{y=0}^{w_{\rho}}\rho(x,y) - \hat{\rho}(x,y),
\end{aligned}
\end{equation}
where $\lambda$ is a weighting factor that allows individual tuning depending on the base model. 
Integrating the density map into the loss function maintains the end-to-end learning characteristic of transformer models.

\section{EXPERIMENTS}
\subsection{Experimental Setup}
\textbf{Datasets:}
We evaluate our presented method on the STCrowd dataset \cite{congSTCrowdMultimodalDataset2022}, which focuses on pedestrians in crowded scenes.
It contains 10,981 images in a total of 84 scenes.  
The scenes are split into 31 for training, 26 for validation, and 27 for testing.
The annotations include 2D boxes, 3D cuboids, and occlusion data for all persons.
Note that the test split annotations are not available.

Although STCrowd offers 2D and 3D box labels, they cannot be completely matched. 
Instead, we project the 3D boxes to the image and approximate the respective 2D box using the outer bounds of the projected boxes.

\begin{table}[!t]
    \caption{Performance comparison on STCrowd 2D detection. * indicates results taken from \cite{liuGigaHumanDetExploringFullBody2024}, which does not report MR\textsuperscript{-2}. Other results are generated by us.}
    \label{table:2d_results}
    \begin{tabularx}{\linewidth}{cllcc}
        \toprule
        & \textbf{Method} & \textbf{Backbone} & \textbf{AP} $\uparrow$ & \textbf{MR\textsuperscript{-2}} $\downarrow$ \\
        \midrule
        \multirow{5}{*}{\rotatebox[origin=c]{90}{\shortstack[c]{\textit{box}\\\textit{based}}}} 
        & Faster R-CNN* \cite{renFasterRCNNRealTime2016} & ResNet-101 & 87.8 & - \\
        & Cascade R-CNN* \cite{caiCascadeRCNNDelving2018} & ResNet-101 & 88.7 & - \\
        & RetinaNet* \cite{linFocalLossDense2017} & ResNet-101 & 88.8 & - \\
        & Pedestron* \cite{hasanGeneralizablePedestrianDetection2021} & HRNet & 90.5 & - \\
        & CrowdDet \cite{chuDetectionCrowdedScenes2020} & ResNet-50 & 89.6 & \textbf{30.7} \\
        \midrule
        \multirow{3}{*}{\rotatebox[origin=c]{90}{\shortstack[c]{\textit{keypoint}\\\textit{based}}}} 
        & CenterNet* \cite{duanCenterNetKeypointTriplets2019} & Hourglass-104 & 84.2 & - \\
        & CentripetalNet* \cite{dongCentripetalNetPursuingHighQuality2020} & Hourglass-104 & 88.7 &  -\\
        & GigaHumanDet* \cite{liuGigaHumanDetExploringFullBody2024} & Hourglass-104 & \underline{90.7} & - \\ 
        \midrule
        \multirow{4}{*}{\rotatebox[origin=c]{90}{\shortstack[c]{\textit{query}\\\textit{based}}}} 
        & DINO* \cite{zhang2023dino} & ResNet-101 & 90.6 & - \\
        & Def-DETR \cite{zhuDeformableDETRDeformable2021} & ResNet-50 & 89.6 & 39.5 \\
        & Iter-Def-DETR \cite{zhengProgressiveEndtoEndObject2022}  & ResNet-50 & 90.5 & 37.7 \\ 
        & DDQ \cite{zhang2023dense} & ResNet-50 & 90.0 & 38.5 \\
        \cmidrule(lr{-0.6em}){2-5}
        & \textbf{CQ2D} & ResNet-50 & \textbf{91.4} & \underline{33.2}\\
        \bottomrule
    \end{tabularx}
\end{table}
\begingroup
\setlength{\tabcolsep}{5pt}
\begin{table}[!t]
    \caption{Performance comparison on STCrowd 3D detection. Results of methods marked with $\dag$ are taken from \cite{congSTCrowdMultimodalDataset2022}. Others are implemented by us.}
    \label{table:3d_results}
    \begin{center}
        \begin{tabularx}{\linewidth}{llcccc}
            \toprule
            \textbf{Method} & \textbf{Backbone} & \textbf{mAP} $\uparrow$ & \textbf{AR\textsubscript{0}} $\uparrow$ & \textbf{AR\textsubscript{1}} $\uparrow$ & \textbf{AR\textsubscript{2}} $\uparrow$ \\
            \midrule
                CentertNet$\dag$ \cite{duanCenterNetKeypointTriplets2019} & ResNet-101 & 20.3 & 47.8 & 36.1 & 27.3 \\
                CentertNet$\dag$ \cite{duanCenterNetKeypointTriplets2019} & DLA34 & 23.6 & 57.8 & 45.1 & 34.9\\
                MonoDETR \cite{zhangMonoDETRDepthguidedTransformer} & ResNet-50 & 48.5 & 71.2 & 62.3 & 54.0 \\
                MonoFlex \cite{zhangObjectsAreDifferent2021} &  DLA-34  & \underline{49.5} & \textbf{74.6} & \textbf{67.3} & \underline{57.3}  \\
                GUPNet \cite{lu2021geometry} &  DLA-34  & 40.2 & 62.6 & 53.5 & 44.7 \\
            \midrule
                \textbf{CQ3D} & ResNet-50 & \textbf{52.7} & \underline{74.4} & \underline{65.7} & \textbf{57.6}\\
            \bottomrule
        \end{tabularx}
    \end{center}
\end{table}
\endgroup

\textbf{Metrics:}
The evaluation of the 2D detection models is based on the widely used \ac{AP} and Miss Rate (MR\textsuperscript{-2}) \cite{shaoCrowdHumanBenchmarkDetecting2018}.

For 3D evaluation, we use the metrics presented by STCrowd \cite{congSTCrowdMultimodalDataset2022}. 
First, the \ac{mAP}, based on the \ac{AP} definition of \cite{caesarNuScenesMultimodalDataset2020}, is employed. 
The 3D center distance serves as the matching criterion.
Three different distance thresholds $D=\{0.25, 0.5, 1\}$ are used and averaged to calculate the final \ac{mAP}. 
Second, the \ac{AR} metric is introduced, which captures the model's recall performance for different occlusion levels.
The \ac{AR} is calculated for the occlusion labels 0 (no occlusion), 1 (no more than half the body), and 2 (over half the body). 
For occlusion label $i$ the $AR_i$ is determined by:
\begin{equation}
    AR_i=\frac{1}{|D|}\sum_{d \in D} Recall_{i,d}, i \in \{0, 1, 2\}.    
\end{equation}

\begin{table}[t]
    \caption{Impact of Density guidance on 3D Object Detection tested on STCrowd. The notations @1 and @3 indicate in which decoder layer the density cross-attention is utilized.}
    \label{table:3d_effect}
    \begin{center}
        \begin{tabularx}{\linewidth}{lcccc}
            \toprule
            \textbf{Architecture} & \textbf{mAP} $\uparrow$ & \textbf{AR\textsubscript{0}} $\uparrow$ & \textbf{AR\textsubscript{1}} $\uparrow$ & \textbf{AR\textsubscript{2}} $\uparrow$\\
            \midrule
                \textbf{CQ3D} & \textbf{52.7} & \textbf{74.4} & \textbf{65.7} & \textbf{57.6} \\
                \midrule
                w/o Density Transformer &  48.5 & 71.2 & 62.3 & 54.0 \\
                w/o Density-guided Embed & 49.9 & 73.5 & \underline{64.7} &  \underline{56.8} \\
                Density Transformer @1 & 50.7 & 72.1 & 63.5 & 54.1 \\
                Density Transformer @3 & \underline{52.6} & \underline{74.0} & 64.6 & 55.8 \\
            \bottomrule
        \end{tabularx}
    \end{center}
\end{table}
\begin{table}[t]
    \caption{Design choice of CQ3D based on the components of self-attention (SA), density cross-attention (\textbf{DS}), depth cross-attention (D), and visual cross-attention (V) tested on STCrowd.}
    \label{table:3d_design}
    \begin{center}
        \begin{tabularx}{\linewidth}{lcccc}
            \toprule
            \textbf{Architecture} & \textbf{mAP} $\uparrow$ & \textbf{AR\textsubscript{0}} $\uparrow$ & \textbf{AR\textsubscript{1}} $\uparrow$ & \textbf{AR\textsubscript{2}} $\uparrow$ \\
            \midrule
                \textbf{DS} $\rightarrow$ SA $\rightarrow$ D $\rightarrow$ SA $\rightarrow$ V &  \textbf{52.7} & \textbf{74.4} & \textbf{65.7} & \textbf{57.6} \\
                D $\rightarrow$ SA $\rightarrow$ \textbf{DS} $\rightarrow$ SA $\rightarrow$ V  & 51.4 & 73.5 & 64.4 & 57.5 \\
            \bottomrule
        \end{tabularx}
    \end{center}
\end{table}
\textbf{Training:}
Our implementation of CQ2D is based on mmdetection \cite{mmdetection}, and the implementation of CQ3D extends the original code from MonoDETR \cite{zhangMonoDETRDepthguidedTransformer}. Following \cite{zhengProgressiveEndtoEndObject2022}, we increase the number of queries to 1000.
If not stated otherwise, we set the loss weight $\lambda$ to 1 for the 2D model and to 0.5 for the 3D model.
Both models use \text{$n_{\text{bins}}=121$} and \text{$\sigma_{x/y}= \frac{1}{3}w_{\text{bbox}}/h_{\text{bbox}}$}. 
The density value range for the embedding is set to \text{$[\rho_{\text{min}}=0, \rho_{\text{max}}=3]$}. 
The lower bound $\rho_{\text{min}}$ is always set to zero to cover empty regions, while the upper bound is set based on the dataset characteristics. 
For STCrowd, we found an average number of $0.02$ overlapping triplets per image, justifying our upper density limit.
We define overlapping triplets by a shared $\text{\acs{IoU}}\geq0.3$.

In our experiments, both models were trained on a single NVIDIA A100 GPU with a batch size of 8.
For the 2D training, we employed a learning rate of $5 \times 10^{-4}$ and trained for 40 epochs. 
The learning rate was reduced by a factor of 2 at epochs 25 and 35.
Regarding the 3D training, we adhered to the settings of MonoDETR.

\subsection{Results}
We report 2D detection results on the STCrowd dataset in Table \ref{table:2d_results}.
DDAD \cite{tangDDADDetachableCrowd2022} is excluded from the baselines, as we were unable to reproduce meaningful detection performance with our local training.
Compared to all other detection methods, our proposed \textbf{CQ2D} achieves the best \ac{AP} with $91.4\%$, outperforming the current best approach GigaHumanDet \cite{liuGigaHumanDetExploringFullBody2024}, by $0.7$ points.
Extending the baseline Deformable DETR \cite{zhuDeformableDETRDeformable2021} model by our proposed \ac{CQ} module improves the \ac{AP} by $1.8$ and the MR\textsuperscript{-2} by $6.3$ points. 
Considering the MR\textsuperscript{-2} metric, our model ranks second, with a $2.5$ point difference compared to CrowdDet \cite{chuDetectionCrowdedScenes2020}, which achieves $30.7\%$.
In general, we observed a less stable MR\textsuperscript{-2} performance during training and worse final performance for all query-based approaches compared to the box-based approach.

Next, we discuss the experiments based on the STCrowd 3D detection task, which are presented in Table \ref{table:3d_results}. 
The methods taken from STCrowd \cite{congSTCrowdMultimodalDataset2022} (see $\dag$) stand out due to their poor performance. 
Investigating the difference, we found that the evaluation setting in \cite{congSTCrowdMultimodalDataset2022} includes all objects in front of the camera within a predefined rectangular area.
Therefore, objects outside the \ac{FoV} of the camera are considered during evaluation, which cannot be detected by a camera-only approach.
Filtering based on the camera \ac{FoV} reduces the number of ground truth instances by roughly 30\%. 
When evaluating MonoFlex without and with the additional instances, the \ac{mAP} drops from 49.5\% to 30.6\%.
This explains the large delta between the reported numbers in \cite{congSTCrowdMultimodalDataset2022} and our experiments.
Still, these numbers indicate a relative improvement of $3.3$ \ac{mAP} points when replacing the ResNet-101 \cite{heDeepResidualLearning2016} backbone with the more advanced DLA-34 \cite{yuDeepLayerAggregation2018}.
Our \textbf{CQ3D} architecture outperforms the baseline MonoDETR \cite{zhangMonoDETRDepthguidedTransformer} for all reported metrics. 
The \ac{mAP} improves by $4.2$, the $AR_0$ by $3.2$, the $AR_1$ by $3.4$, and the $AR_2$ by $3.6$.
Therefore, our method improves the detection performance using only 2D boxes, which are based on the projected 3D bounding boxes, as a ground truth for the density task.
Furthermore, our method surpasses the \ac{mAP} of any other tested method by at least $3.2$ points, even though it uses the simpler ResNet-50 backbone. 
Reviewing the AR, we exhibit lower scores than MonoFlex on fully visible $(AR_0)$ and lightly occluded $(AR_1)$ instances, though we perform best on highly occluded instances $(AR_2)$.

\begin{figure*}[t]
    \hfill
    \begin{minipage}[b]{0.24\textwidth}
        \centering
        \includegraphics[width=\textwidth]{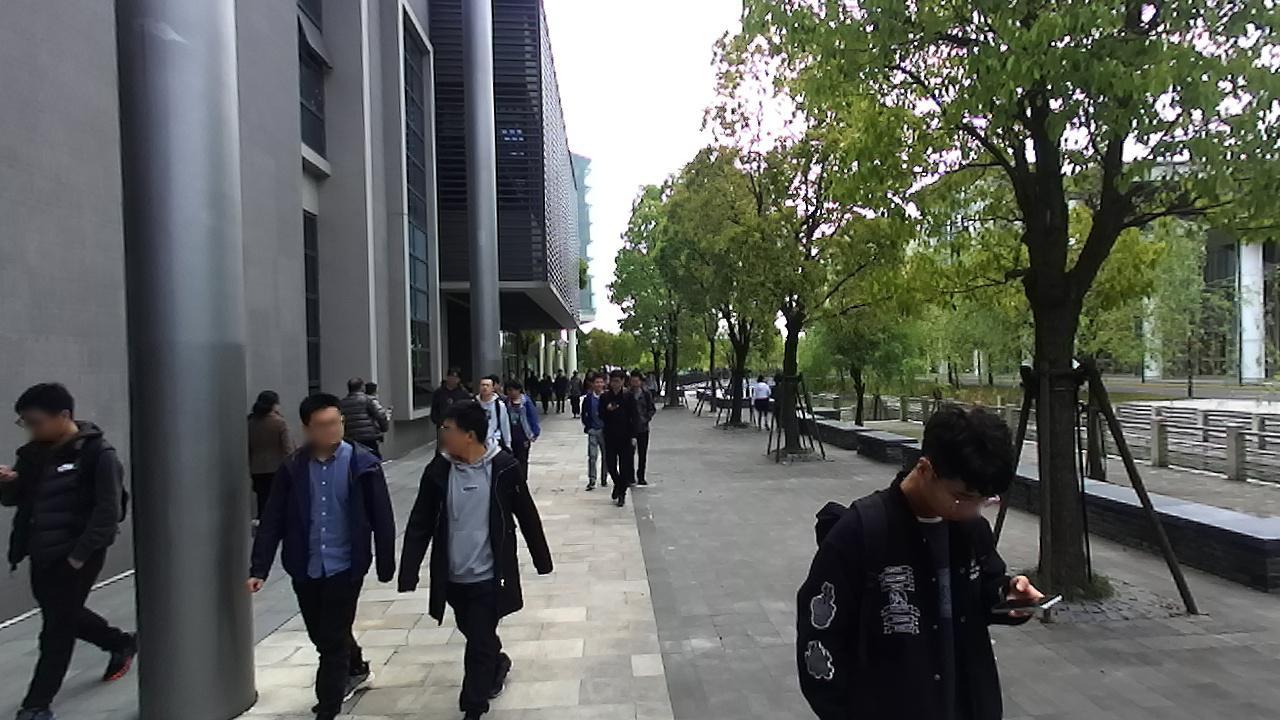}
    \end{minipage}
    \hfill
    \begin{minipage}[b]{0.24\textwidth}
        \centering
        \includegraphics[width=\textwidth]{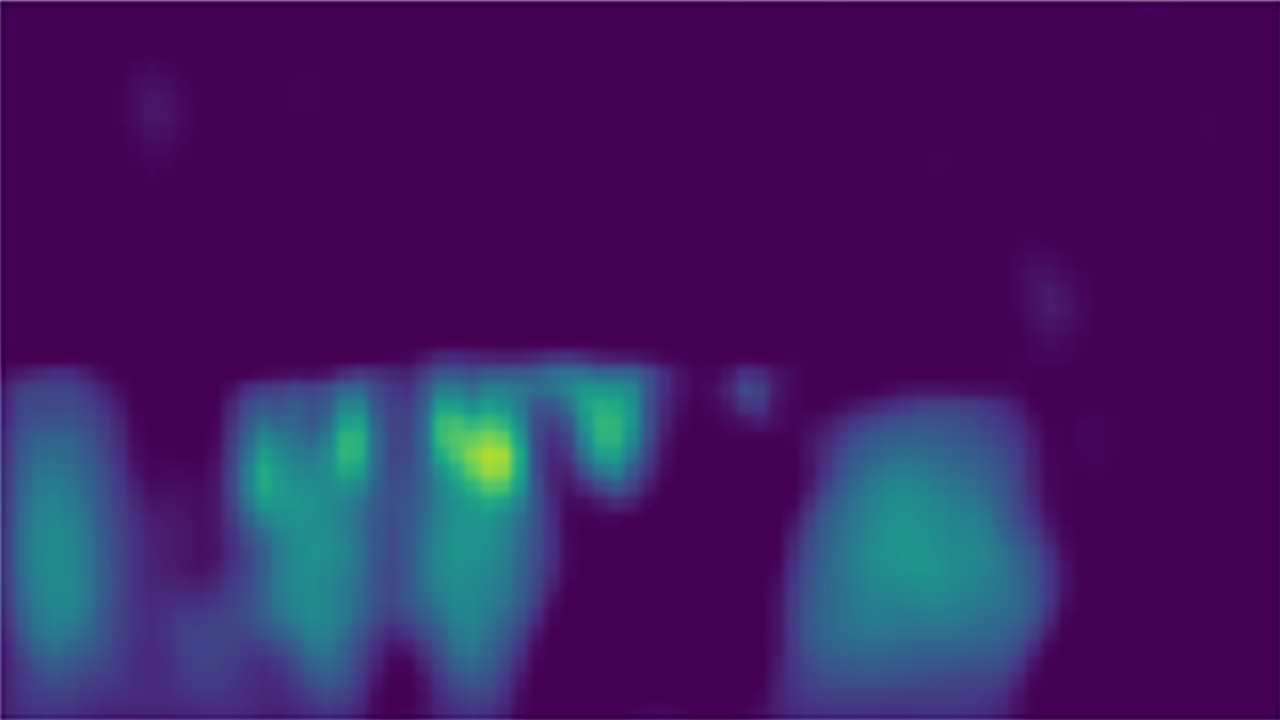}
    \end{minipage}
    \hfill
    \begin{minipage}[b]{0.24\textwidth}
        \centering
        \includegraphics[width=\textwidth]{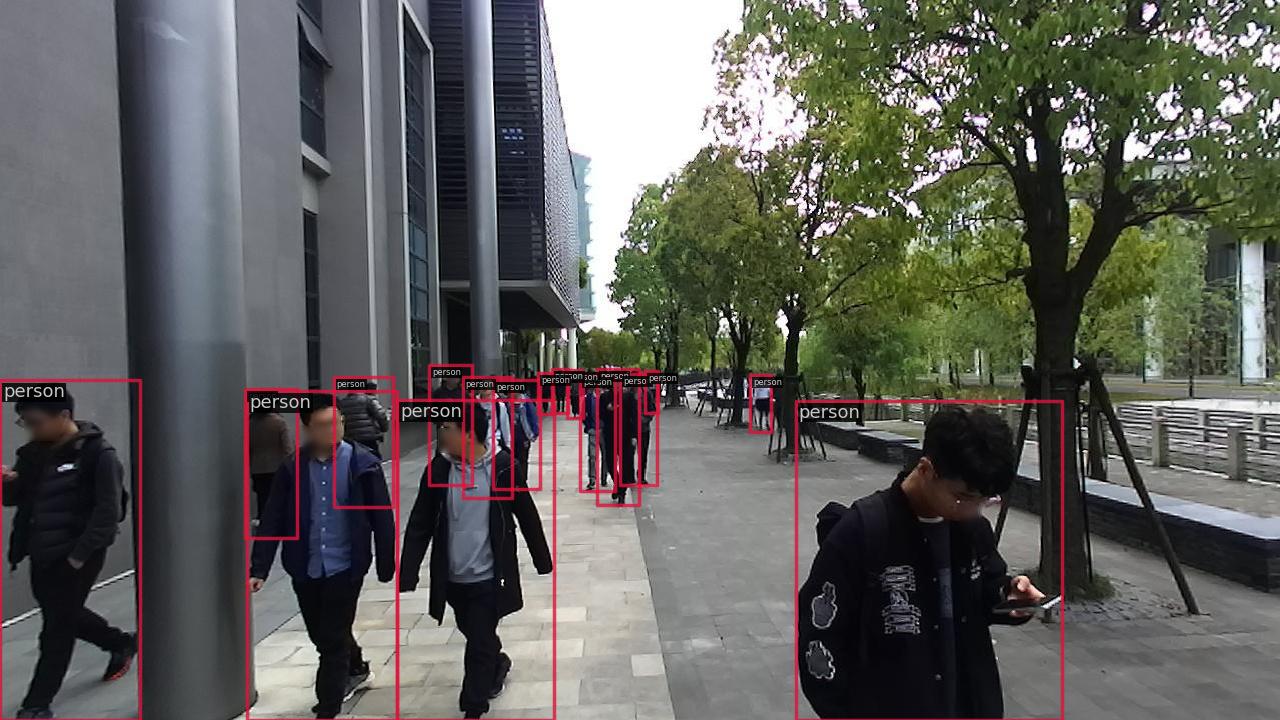}
    \end{minipage}
    \hfill
    \begin{minipage}[b]{0.24\textwidth}
        \centering
        \includegraphics[width=\textwidth]{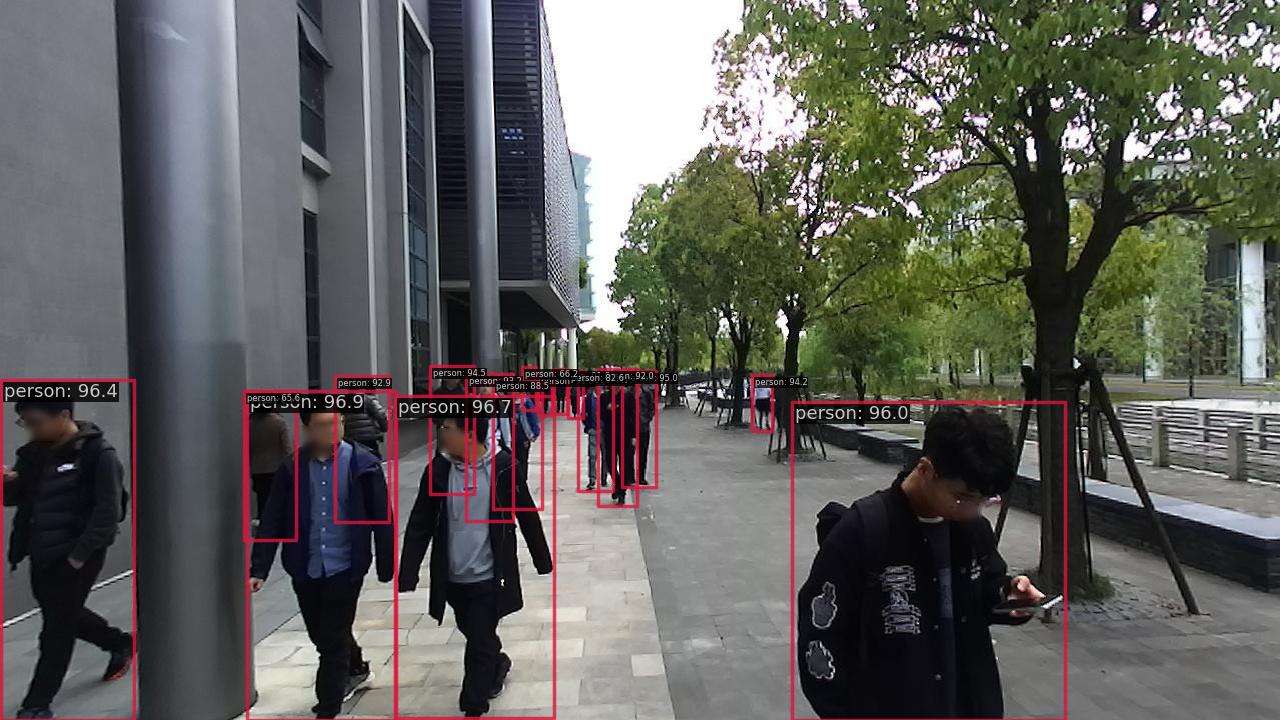}
    \end{minipage}
    \hfill
    
    \vspace{0.5em} %

    \hfill
    \begin{minipage}[b]{0.24\textwidth}
        \centering
        \includegraphics[width=\textwidth]{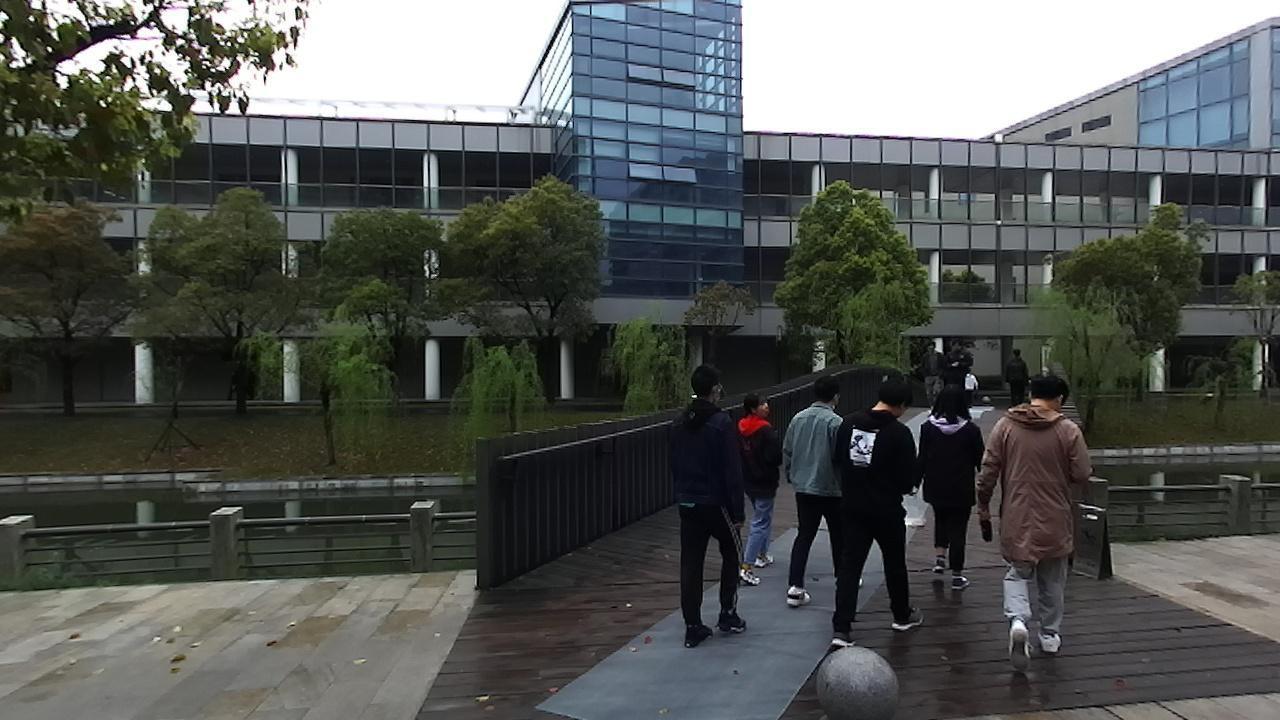}
    \end{minipage}
    \hfill
    \begin{minipage}[b]{0.24\textwidth}
        \centering
        \includegraphics[width=\textwidth]{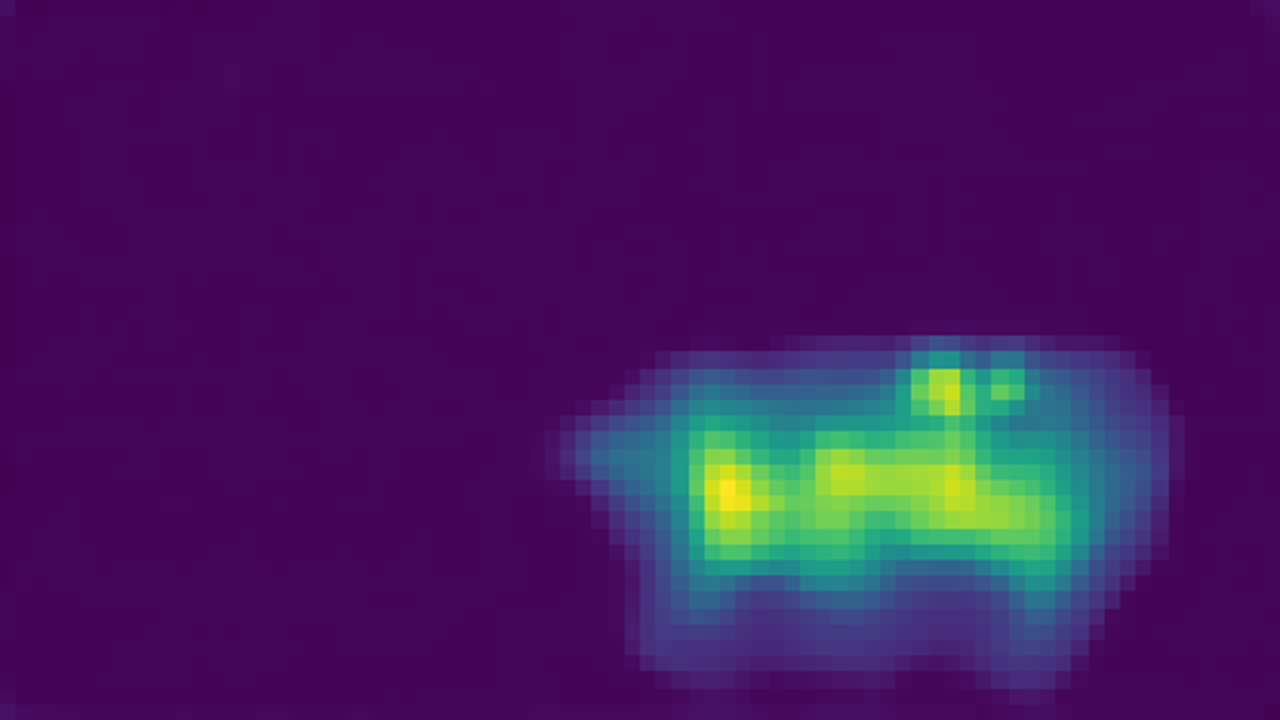}
    \end{minipage}
    \hfill
    \begin{minipage}[b]{0.24\textwidth}
        \centering
        \includegraphics[width=\textwidth]{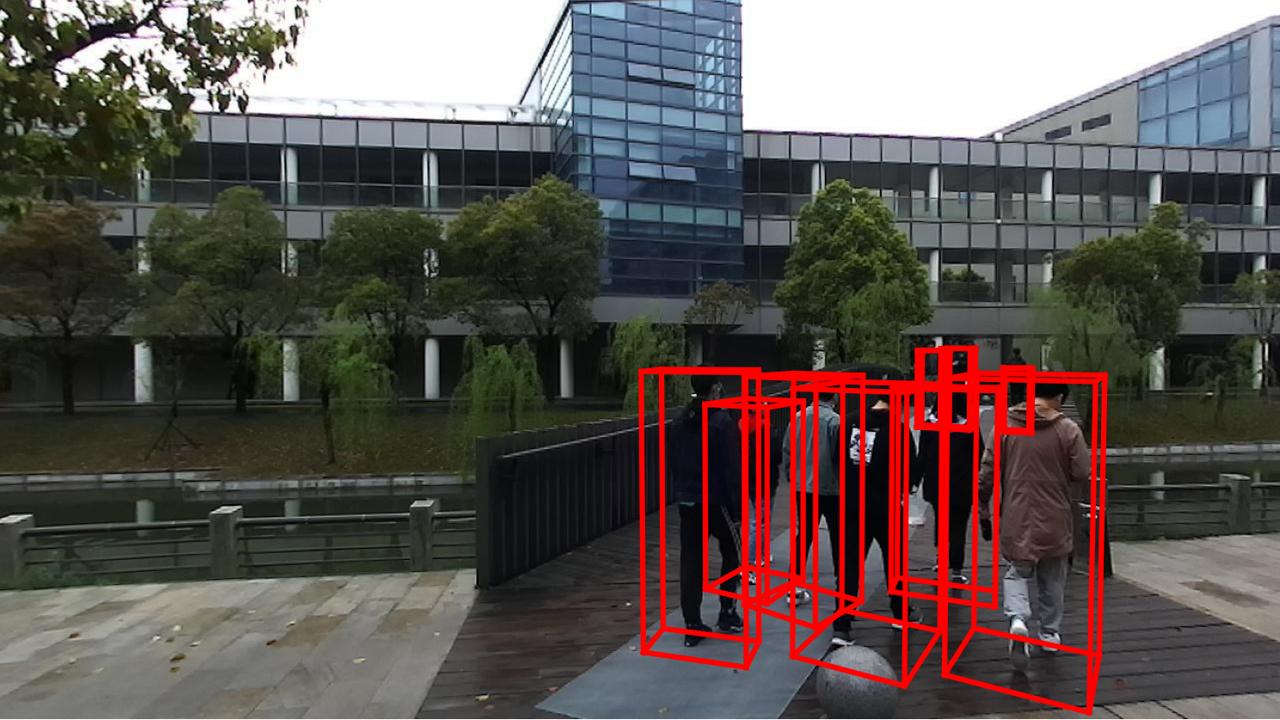}
    \end{minipage}
    \hfill
    \begin{minipage}[b]{0.24\textwidth}
        \centering
        \includegraphics[width=\textwidth]{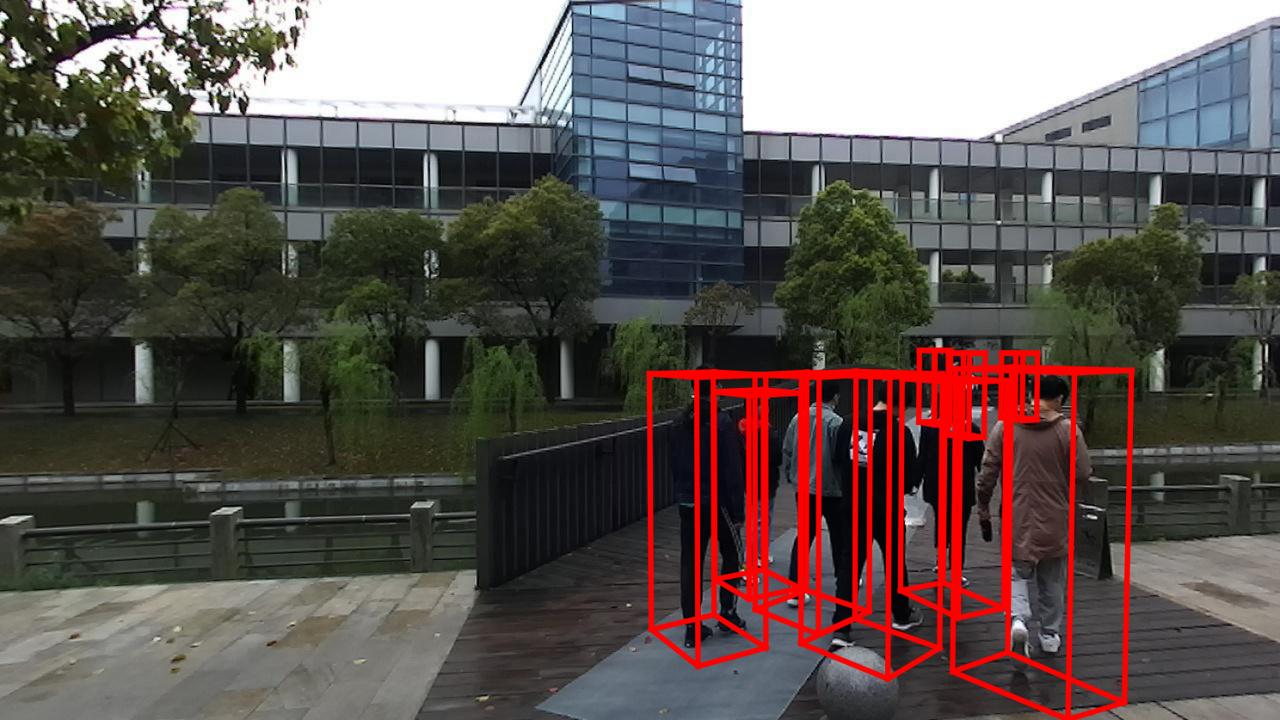}
    \end{minipage}
    \hfill
    \caption{Qualitative results of CQ2D in the upper row and CQ3D in the lower row tested on STCrowd. From left to right, the images first show the original, the predicted density map, the ground truth objects, and the prediction objects. The numbers above each prediction indicate the confidence score. These are omitted for the 3D case for better readability. The predicted density maps effectively highlight dense regions. While artifacts are visible in the upper density map, they do not result in false positives, indicating no negative impact on model performance. The 2D detections largely overlap with the ground truth, showing good detection performance. In 3D detections, the model successfully predicts the object centers and dimensions close to the ground truth, ensuring no objects are missed. However, occlusions make it challenging to predict the yaw angle, resulting in visible deviations.}
    \label{fig:qual}
\end{figure*}

In addition to the quantitative findings, we also provide qualitative results of both CQ2D and CQ3D in Figure \ref{fig:qual}.
\subsection{Ablations}

\begin{table}[!t]
    \caption{Comparison of architecture configurations tested on STCrowd 2D detection. The final row represents our final and best configuration.}
    \label{table:2d_embed_map}
    \begin{center}
        \begin{tabular}{lcccc}
            \toprule
            \textbf{Embedding} & $\boldsymbol{d}$ & $\boldsymbol{n_{\text{bins}}}$ &\textbf{AP} $\uparrow$ & \textbf{MR\textsuperscript{-2}} $\downarrow$ \\
            \midrule
            Linear Increasing & $\sfrac{1}{3}$ & 121 & 91.0 & 35.0 \\
            Linear Decreasing & $\sfrac{1}{3}$ & 121 & 90.3 & 38.4 \\
            \midrule
            Uniform & $\sfrac{1}{3}$ & 41 & 90.1 & 36.8 \\
            Uniform & $\sfrac{1}{3}$ & 21 & 89.2 & 37.4 \\
            \midrule
            Uniform & $\sfrac{1}{2}$ & 121 & 91.1 & 34.3 \\
            Uniform & $\sfrac{1}{4}$ & 121 & 90.2 & 36.2 \\
            Uniform & $\sfrac{1}{6}$ & 121 & 90.7 & 35.1 \\
            \midrule
            \rowcolor[HTML]{EFEFEF}
            Uniform & $\sfrac{1}{3}$ & 121 & \textbf{91.4} & \textbf{33.2} \\
            \bottomrule
        \end{tabular}
    \end{center}
\end{table}
Subsequently, we investigate the impact of the presented components on the total performance.
Given the notably larger improvements in 3D detection and increased task complexity relative to 2D, the efficiency analysis focuses on CQ3D to provide a clearer and more impactful evaluation of our method's advancements.
Additionally, we investigate the optimal placement of our CQ module within the selected base 3D detector, MonoDETR.
The general density configurations are assessed using the 2D task to provide broader insights. 
Finally, we assess the computational overhead using CQ3D.

\textbf{Effectiveness of Density-guided Transformer:}
Table \ref{table:3d_effect} demonstrates the impact of introducing density guidance.
The baseline \lq w/o Density Transformer\rq, which is equivalent to MonoDETR, performs worse than any other architecture that contains density supervision.
Adding solely the extra head to predict the density map (\lq w/o Density-guided Embed\rq) improves all metrics, especially the ARs.
Meanwhile, integrating the density-guided transformer into selected layers predominantly improves the mAP. 
Moreover, the density guidance is more effective in the last layer than in the first one. 
However, adding the density-guided transformer to every layer achieves the best overall performance.

\begin{table}[t]
    \caption{Evaluation of CQ2D combined with DDQ\cite{zhang2023dense} on STCrowd 2D, referred to as CQ++, with different density loss weighting $\lambda$.}
    \label{table:cq2d_ddq}
    \begin{center}
        \begin{tabular}{lccc}
            \toprule
            \textbf{Architecture} & $\boldsymbol{\lambda}$ & \textbf{AP} $\uparrow$ & \textbf{MR\textsuperscript{-2}} $\downarrow$ \\
            \midrule
            DDQ \cite{zhang2023dense} & - & 90.0 & 38.5 \\
            \midrule
            \textbf{CQ2D++} & 1 & 91.6 & 32.2 \\
            \textbf{CQ2D++} & 5 & \textbf{92.3} & \textbf{31.4} \\       
            \bottomrule
        \end{tabular}
    \end{center}
\end{table}

\textbf{3D Decoder Structure:}
We conduct two experiments to determine the optimal placement for our CQ module in the decoder.
We always use self-attention after the density and depth cross-attention to allow information exchange between the queries. 
Following \cite{zhangMonoDETRDepthguidedTransformer}, the visual cross-attention is always placed last.
Table \ref{table:3d_design} shows the advantage of placing the density cross-attention before the depth cross-attention.

\textbf{Embedding Mapping:}
The top part in Table \ref{table:2d_embed_map} compares the effect of using different embedding mappings.
The mapping defines how the density value range is mapped to embedding bins. 
We test three mapping options: 
A \textit{uniform} mapping, with evenly distributed bins, a \textit{linear increasing discretization}, and a \textit{linear decreasing discretization}.
The latter two cover the value ranges with bins that get smaller/larger as the density values increase, respectively.
I.e., there are more finely grained embeddings of higher/lower density values. 
To our surprise, the uniform embedding, which is also the simplest, performed best. 
This indicates a uniform importance across the density value range.

\textbf{Embedding Bin Count:}
The center part of Table \ref{table:2d_embed_map} shows the impact of the number of bins used for the density embedding.
We select a variety of three bin counts, ranging from coarse to fine. 
The bin counts we evaluate are \mbox{$n_{\text{bins}} = \{21, 41, 121\}$} corresponding to the bin widths 0.15, 0.075, and 0.025, respectively. 
The highest resolution with $121$ bins works best, which indicates that a distinction between the value ranges is important. 

\textbf{Density Map:}
In the bottom part of Table \ref{table:2d_embed_map}, we investigate the impact of the chosen $\sigma$ values for generating the target density map by evaluating different scaling factors.
We have chosen four different settings \mbox{$d = \{\frac{1}{2}, \frac{1}{3}, \frac{1}{4}, \frac{1}{6}\}$} to cover a wide range of values in the context of 2D detection.
By selecting $d=\frac{1}{3}$, we achieve optimal performance, which results in \text{$\sigma_{x}=\frac{1}{3}w_{\text{bbox}}$} and \text{$\sigma_{y}=\frac{1}{3}h_{\text{bbox}}$}.

\begin{table}[t]
    \caption{2D detection performance on Crowdhuman dataset.}
    \label{tab:crowdhuman}
    \begin{center}
        \begin{tabular}{lccc}
            \toprule
            \textbf{Method} & \textbf{Backbone} & \textbf{AP} $\uparrow$ &  \textbf{MR\textsuperscript{-2}} $\downarrow$ \\
            \midrule
            CrowdDet \cite{chuDetectionCrowdedScenes2020} & ResNet-50 & 90.7 & 41.4 \\
            Iter-Def-DETR \cite{zhengProgressiveEndtoEndObject2022} & ResNet-50 & 92.1 & 41.5 \\
            UniHCP \cite{ci2023unihcp} & ViT & 92.5 & 41.6 \\
            DDQ \cite{zhang2023dense} & ResNet-50 & 93.8 & 39.7 \\
            \midrule
            \textbf{CQ2D++} & ResNet-50 & \textbf{94.0} & \textbf{39.0} \\
            \bottomrule
        \end{tabular}
    \end{center}
\end{table}

\begin{table}
    \centering
    \begin{tabular}{lll}
        \toprule
         \textbf{Method} & \textbf{FPS} & \textbf{Params} \\
         \midrule
         Baseline (MonoDETR) & 15.6 & 37,954,768 \\
         CQ3D & 13.1 \red{(-16.0\%)} & 42,855,121 \red{(+12.9\%)} \\
         \bottomrule
    \end{tabular}
    \caption{Impact of our method on the runtime efficiency and parameter count based on CQ3D. FPS are measured on a single NVIDIA 2080Ti using a batch size of 1 of STCrowd data.}
    \label{tab:compute}
\end{table}

\subsection{Integration into Crowd-Focused Detector}\label{subsection:crowd_focus}
Besides the integration into general detectors, we investigate the application in crowd-focused detectors. 
Due to the lack of suitable 3D detectors, we prioritize 2D detectors and select DDQ \cite{zhang2023dense}. 
As shown in Table \ref{table:cq2d_ddq}, DDQ already performs well on STCrowd 2D without any modifications.
By adding our \ac{CQ} module to DDQ, referred to as CQ2D++, the model outperforms all other methods presented in Table \ref{table:2d_results} as well as CQ2D in AP. 
Additionally, CQ2D++ closes the gap of the MR\textsuperscript{-2} to 0.7 points compared to \cite{chuDetectionCrowdedScenes2020}.
Optimal results were achieved by tuning the density loss more aggressively, that is, we increased $\lambda$, to account for the model's pre-existing crowd focus.
In the end, CQ2D++ improves the AP by 2.3 and the MR\textsuperscript{-2} by 7.1 points. 
All experiments are performed using the original DDQ configuration.

\subsection{Dataset Transfer}
A major challenge in our study is the lack of suitable 3D datasets. 
While there are multiple monocular 3D datasets such as KITTI \cite{geigerAreWeReady2012}, they rarely exhibit a significant number of pedestrians, especially not in crowded scenarios. 
However, in 2D detection, there are multiple viable datasets.
We choose the widely used CrowdHuman \cite{shaoCrowdHumanBenchmarkDetecting2018} dataset due to its high object density.
To ensure competitiveness, we select our best model, CQ2D++ (cf. Section \ref{subsection:crowd_focus}), and evaluate it on the challenging CrowdHuman dataset (see results in Table \ref{tab:crowdhuman}). 
CQ2D++ mainly improves MR\textsuperscript{-2} and outperforms all other methods.
Even though results on CrowdHuman have shown to converge over the last publications, we are still able to improve nearly one point in MR\textsuperscript{-2} using our easy-to-integrate method.
We kept the optimal settings of CQ2D++ from the previous experiments to highlight easy transferability. 
Additional fine-tuning might enhance results further.

\subsection{Computation Efficiency}\label{subsection:compute}
As shown in Table \ref{tab:compute}, CQ3D introduces a $16.0\%$ runtime increase, with two-thirds attributed to the self-attention in the \ac{CQ} module. 
The parameter count increases by just $12.9\%$, highlighting our method's computational efficiency.

\section{CONCLUSION}
We presented a general method called \acl{CQ}, which uses object density information to guide queries and enhance both general and crowd detectors. 
The method integrates easily into preexisting networks, as long as a query-based decoder is present. 
Integrating \ac{CQ} into selected base models results in the architectures CQ2D and CQ3D.
Both architectures demonstrate superior performance compared to non-extended versions and mostly outperform or match state-of-the-art methods. This improvement is not limited to the selected base models, as shown by the clear improvement when integrated into the crowd-based model DDQ \cite{zhang2023dense}.
Extensive ablation studies validate the effectiveness of our method. To achieve optimal results, the parameters should be adapted to the selected base model. 
We hope to inspire others to treat both 2D and 3D crowd detection as a more unified topic in the future.
Future work will explore our method’s generalizability by including more networks and datasets, and by extending it to LiDAR-based detectors.

\section*{ACKNOWLEDGMENT}
This work is a result of the joint research project STADT:up (19A22006O). The project is supported by the German Federal Ministry for Economic Affairs and Climate Action (BMWK), based on a decision of the German Bundestag. The author is solely responsible for the content of this publication.


\end{document}